# Comparisons among different stochastic selection of activation layers for convolutional neural networks for healthcare

**Loris Nanni [1]\*, Alessandra Lumini[2], Stefano Ghidoni[1] and Gianluca Maguolo[1]**

1. DEI, University of Padua, viale Gradenigo 6, Padua, Italy
2. DISI, Università di Bologna, Via dell'università 50, 47521 Cesena, Italy; alessandra.lumini@unibo.it
\* Correspondence: loris.nanni@unipd.it;

**Abstract:** Classification of biological images is an important task with crucial application in many fields, such as cell phenotypes recognition, detection of cell organelles and histopathological classification, and it might help in early medical diagnosis, allowing automatic disease classification without the need of a human expert. In this paper we classify biomedical images using ensembles of neural networks. We create this ensemble using a ResNet50 architecture and modifying its activation layers by substituting ReLUs with other functions. We select our activations among the following ones: ReLU, leaky ReLU, Parametric ReLU, ELU, Adaptive Piecewice Linear Unit, S-Shaped ReLU, Swish , Mish, Mexican Linear Unit, Gaussian Linear Unit, Parametric Deformable Linear Unit, Soft Root Sign (SRS) and others.

We tested multiple ensembles obtained using the sum rule. The sum rule is a method to combine the results of multiple networks and it consists in averaging the output probability vectors of every network. Our test can be divided into two categories.

• We substituted every ReLU in the original network with all the activations that we named earlier. In this way, we created many different networks, each of which has different activation functions. However, all the activation layers in a network have the same activation.

• We substituted every ReLU in the original network with a randomly selected activation. In this case, however, every activation layer might contain a different activation function. We repeated this process multiple times, since its stochasticity allows us to get a different network every time. We tried to use different pools of activation functions.

As a baseline, we used an ensemble of neural networks that only use ReLU activations. We tested our networks on several small and medium sized biomedical image datasets. Our results prove that our best ensemble obtains a better performance than the ones of the naive approaches. In order to encourage the reproducibility of this work, the MATLAB code of all the experiments will be shared at https://github.com/LorisNanni.

## 1. Introduction

Deep Learning has been on the stage in the last years because it was able to set new levels of performance in a variety of fields like image processing and natural language processing [1]. Like other neural networks, deep networks make use of non-linear activation functions (e.g., tanh, ReLU) that have a strong influence on the performance that can be achieved, as they provide to the network the ability of approximating an arbitrarily complex function.

Given the effect on the network performance, in this paper we compare a large number of activation functions used in Convolutional Neural Networks (CNNs) for image classification tasks on multiple and diverse datasets. Our backbone network is ResNet50 [3] and we modify it by changing its original activation functions with other ones proposed in the literature. This process returns a large number of different networks that we use to create ensemble of CNNs.

At first, we introduce the activation functions that we shall substitute in the original network. At that point, we introduce our approach to randomly substitute all the activations in the original network. We substitute them with new ones chosen among the ones that we introduced earlier. The key idea is that the selections of the activations are independent in every layer. Hence, this procedure always yields a different network, providing a large number of partially independent classifiers. The classifiers can be fused together to create an ensemble of CNNs. Ensembles of classifiers have been proven to be more effective than stand-alone CNNs due to the instability of their training [4], hence our methods allows to boost the performance of the single CNNs.

We evaluate our approach on several medical dataset following our protocol that includes a fine-tuning of each model considered on each dataset. Medical images are a good test for CNNs because they show a variety of low- and mid-level patterns of different dimensions [5]. CNNs have already demonstrated to be effective analyzing such kind of images in a variety of applications like carcinoma and melanoma detection [6], classification of subcellular and stem cell images [7], thyroid nodules classification [8], breast cancer detection [9].

Our experiments show that the proposed ensembles work well in all the tested problems gaining state-of-the-art classification performance [10].

The code developed for this work will be available at https://github.com/LorisNanni

## 2. Literature reviews

Deep learning started a new era in several fields, including many computer vision applications like pattern recognition, object detection and many others [11]. Convolutional Neural Networks (CNNs) are widely used for processing images, thanks to their convolutional layers that process image considering local neighborhood. The layered approach [12] demonstrated to be very effective analyzing the low-level and mid-level patterns for image processing. Researchers gradually increased the number of layers, as this was able to improve the representation learned by the models and increase the performance in several domains. This, however, comes at the expense of a need of very large datasets and of an increased risk of overfitting, that pushed towards a large use of data augmentation, regularization techniques and more effective activation functions.

The role of activation functions is sometimes underestimated, but it is central to gain high performance, as it introduces a strong non-linear component. This is a topic focusing the attention of several researchers [13][14]. One of the main advances in this topic is the introduction of the Rectified Linear Unit (ReLU) [15], a piecewise linear function that substituted the sigmoid function because it is extremely fast to compute and offers very good performance. Building on this success, a family of activation functions was developed based on the ReLU function: i) Leaky ReLU [16], that shows a small slope ($\alpha$) for negative inputs; ii) ELU (Exponential Linear Unit) [14], showing an exponential decrease to a limit point $\alpha$ in the negative domain; iii) SELU (Scaled Exponential Linear Unit) [17], a scaled version of ELU by a constant $\lambda$; iv) RLReLU (Randomized Leaky ReLU) [18], based on a non-linear random coefficient.

The functions i-iv just mentioned are static, because they depend on a set of parameters that are chosen and kept constant throughout the whole training process. A more advanced activation function design is capable of adjusting the parameter set at training time, thus adapting the activation function to the training data, making the function dynamic. Dynamic functions usually train a different set of parameter for each layer or, in some cases, for each neuron. This, however, increases the number of learnable parameters, which can increase overfitting. Activation functions belonging to this second group are: v) PReLU (Parametric ReLU) [19], a parametric version of ReLU similar to Leaky ReLU for which the parameter $\alpha$ is not static but, rather, learned; vi) APLU (Adaptive Piecewise Linear Unit) [13], that is a piecewise function determined by a set of parameters that are learned separately for each neuron at training time; vii) Trained Activation Function [20], whose shape is learned by means of a linear regression model – two variants are proposed in [21]: linear sigmoidal activation and adaptive linear sigmoidal activation. According to [22], two of the best performing functions are; viii) Swish, a sigmoid function based on a trainable parameter, and ix) Mexican ReLU (MeLU) [23], a piecewise linear activation function obtained summing PReLU and a number of Mexican hat functions.

In this work we consider mixtures of static and dynamic activation function, with the aim of combining the advantages of both, mitigating their weak points.

## 3. Activation Functions

In this paper we consider several different activations. We shall now briefly introduce them. Many of these activations depend on a hyperparameter called $maxInput$, which is used to normalize the activation depending on whether the input is in [0,1] or [0,255].

The first activation is ReLU [15], which is defined as:

$$y_i = f(x_i) = \begin{cases} 0, & x_i < 0 \\ x_i, & x_i \geq 0 \end{cases} \tag{1}$$

and its derivative is given by:

$$\frac{dy_i}{dx_i} = f'(x_i) = \begin{cases} 0, & x_i < 0 \\ 1, & x_i \geq 0 \end{cases} \tag{2}$$

Many variants of ReLU have been proposed in the literature. The first one that we consider is Leaky ReLU [16], which is defined as:

$$y_i = f(x_i) = \begin{cases} ax_i, & x_i < 0 \\ x_i, & x_i \geq 0 \end{cases} \tag{3}$$

where the $a$ is a small and positive number (0.01 in this work). The main strength of Leaky ReLU is that its derivative is always positive:

$$\frac{dy_i}{dx_i} = f'(x_i) = \begin{cases} a, & x_i < 0 \\ 1, & x_i \geq 0 \end{cases} \tag{4}$$

The second variant of ReLU that we consider is Exponential Linear Unit (ELU) [14], which is defined as:

$$y_i = f(x_i) = \begin{cases} a(\exp x_i - 1), & x_i < 0 \\ x_i, & x_i \geq 0 \end{cases} \tag{5}$$

where $a$ is a positive number (1 in this study). ELU has also a positive gradient and it is also continuous:

$$\frac{dy_i}{dx_i} = f'(x_i) = \begin{cases} a \exp(x_i), & x_i < 0 \\ 1, & x_i \geq 0 \end{cases} \tag{6}$$

Parametric ReLU (PReLU) [24] is a learnable variant of Leaky ReLU. It is defined by:

$$y_i = f(x_i) = \begin{cases} a_c x_i, & x_i < 0 \\ x_i, & x_i \geq 0 \end{cases} \tag{7}$$

where $a_c$ are different real numbers, one for each input channel. PReLU is similar to Leaky ReLU, the only difference being that the $a_c$ parameters are learnable. The gradients of PReLU are:

$$\frac{dy_i}{dx_i} = f'(x_i) = \begin{cases} a_c, & x_i < 0 \\ 1, & x_i \geq 0 \end{cases} \text{ and } \frac{dy_i}{da_c} = \begin{cases} x_i, & x_i < 0 \\ 0, & x_i \geq 0 \end{cases} \tag{8}$$

S-Shaped ReLU (SReLU) [25] is the fourth variant or ReLU that we include in this study. It is defined as a piecewise linear function:

$$y_i = f(x_i) = \begin{cases} t^l + a^l(x_i - t^l), & x_i < t^l \\ x_i, & t^l \leq x_i \leq t^r \\ t^r + a^r(x_i - t^r), & x_i > t^r \end{cases} \quad (9)$$

SReLU depends on four sets of learnable parameters: $t^l, t^r, a^l$, and $a^r$. In this paper, they are initialized to $a^l = 0, t^l = 0, a^r = 1, t^r = maxInput$. Hence, its initialization makes it equal to ReLU in the first training step. The gradients are given by:

$$\frac{dy_i}{dx_i} = f'(x_i) = \begin{cases} a^l, & x_i < t^l \\ 1, & t^l \leq x_i \leq t^r \\ a^r, & x_i > t^r \end{cases} \quad (10)$$

$$\frac{dy_i}{da^l} = \begin{cases} x_i - t^l, & x_i < t^l \\ 0, & x_i \geq t^l \end{cases}, \quad \text{and} \quad (11)$$

$$\frac{dy_i}{dt^l} = \begin{cases} -a^l, & x_i < t^l \\ 0, & x_i \geq t^l \end{cases} \quad (12)$$

Adaptive Piecewise Linear Unit (APLU) [13] is a piecewise linear function whose slopes and points of non-differentiability are learnable. It is defined as:

$$y_i = \text{ReLU}(x_i) + \sum_{c=1}^{n} a_c \min(0, -x_i + b_c) \quad (13)$$

where $n$ is an hyperparameter defining the number of hinges; $a_c$ and $b_c$ are real numbers, one for each input channel. Its gradients are:

$$\frac{df(x,a)}{da_c} = \begin{cases} -x + b_c, & x < b_c \\ 0, & x \geq b_c \end{cases} \quad \text{and} \quad \frac{df(x,a)}{db_c} = \begin{cases} -a_c, & x < b_c \\ 0, & x \geq b_c \end{cases} \quad (14)$$

We initialized the parameters $a_c$ to zero, and the points of non-differentiability are randomly selected. We also added an $L^2$-penalty of 0.001 to the norm of the parameters $a_c$, as suggested by its creators.

Mexican ReLU (MeLU) [23], is a variant of ReLU derived from the Mexican hat functions. These functions are defined as:

$$\phi^{a,\lambda}(x) = \max(\lambda \cdot maxInput - |x - a \cdot maxInput|, 0) \quad (15)$$

where $a$ and $\lambda$ are real numbers. The output of MeLU is defined as:

$$y_i = \text{MeLU}(x_i) = \text{PReLU}^{c_0}(x_i) + \sum_{j=1}^{k-1} c_j \phi^{\alpha_j, \lambda_j}(x_i) \quad (16)$$

which is the weighted sum of PReLU and $k-1$ Mexican hat functions. The weights in PReLU and of those that multiply the Mexican functions are the learnable parameters. $\alpha_j$ and $\lambda_j$ are fixed parameters chosen recursively. We refer to the original paper for a more detailed explanation of how they are chosen. MeLU is continuous and piecewise differentiable. Besides, MeLU generalize ReLU in the sense, when all the $c_i$ parameters are set to zero, the two functions coincide. This is a useful property because MeLU can be substituted in any ReLU network allowing an efficient transfer learning, if it is properly initialized.

The gradient of MeLU is simply the weighted sum of the gradients of PReLU and those of the Mexican hat functions.

In this work, we used two values of $k$. In our experiments, we call MeLU the implementation where $k=4$ and wMeLU the implementation where $k=8$, where wMeLU stands for wider MeLU.

Gaussian ReLU, also called GaLU [26], is an activation based on so-called Gaussian-like functions:

$$\phi_g^{a,\lambda}(x) = \max{(\lambda \cdot maxInput - |x - a \cdot maxInput|, 0)} +$$
$$+ \min{(|x - a \cdot maxInput - 2\lambda \cdot maxInput| - \lambda \cdot maxInput, 0)} \tag{19}$$

where $a$ and $\lambda$ are real numbers. GaLU is defined as:

$$y_i = GaLU(x_i) = PReLU^{c_0}(x_i) + \sum_{j=1}^{k-1} c_j \, \phi_g^{a_j,\lambda_j}(x_i) \tag{20}$$

which is the homologous of MeLU for Gaussian-type functions. In the experiments we call GaLU the implementation with $k = 4$ and sGaLU (smaller GaLU) the one with $k = 2$.

Parametric Deformable Exponential Linear Unit (PDELU) was introduced in [64] and it is defined as:

$$y_i = f(x_i) = \begin{cases} x_i, & x_i > 0 \\ a_i \cdot \left([1 + (1-t)x_i]_+^{\frac{1}{1-t}} - 1\right), & x_i \leq 0 \end{cases}$$

It has zero mean and, according to its creators, this fasten the training process.

Swish is an activation function introduced in [22]. Its creators used reinforcement learning to assemble different basis functions using sum, multiplication and composition. The output of this learning was a smooth and non-monotonic function defined as:

$$y = f(x) = x \cdot sigmoid(\beta x) = \frac{x}{1 + e^{-\beta x}}$$

where $\beta$ is a parameter that can optionally be learnable. In our tests we initialize it to 1.

Mish is an activation function introduced in [27]. It is defined as

$$y = f(x) = x \cdot tanh(softplus(\alpha x)) = x \cdot tanh(ln(1 + e^{\alpha x}))$$

where $\alpha$ is a learnable parameter.

Soft Root Sign (SRS) is an activation introduced in [28]. It is defined as:

$$y = f(x) = \frac{x}{\frac{x}{\alpha} + e^{-\frac{x}{\beta}}}$$

where $\alpha$ and $\beta$ are learnable and non-negative parameters. SRS is neither monotone nor positive. If the distribution of the input is a standard normal, its shape allows it to have zero mean, which again should enable faster training.

Soft Learnable is a new activation function proposed in [28], which is defined as:

$$y = f(x) = \begin{cases} x, & x > 0 \\ \alpha \cdot ln\left(\frac{1 + e^{\beta x}}{2}\right), & x \leq 0 \end{cases}$$

where $\alpha, \beta$ are positive parameters. We used two different versions of this activation, depending on whether the parameter $\beta$ is fixed (SoftLearnable) or learnable (SoftLearnable2).

In order to avoid any overfitting to the data, we used the same parameter setting that the original authors of each activation suggest in their papers.

## 4. Materials and Methods

In this section we detail both the datasets used for performing experimental evaluation, the backbone architectures and the stochastic methods proposed to design new CNN models and create ensembles.

Experimental evaluation is carried out for medical image classification, performing experiments on several well-known medical datasets which are summarized in Table 1. For each dataset the following information is included: dataset name and reference, a short abbreviation, the number of samples and classes, the size of the images and the testing protocol, which is five-fold cross-validation (5CV) in almost all cases except when expressly specified by the authors of the dataset (ten-fold cross-validation (10CV) for CO and a three fold division for the LAR [34]). Even if datasets include images of different size and aspect ratios, all the images have been resized to the fixed squared size of 224×224 required from the input size of the most known CNN model.

**Table 1**. Summary of the medical image datasets used for image classification: short same (ShortN), name, number of classes (#C), number of samples (#S), image size, testing protocol, reference.

| ShortN | Name | #C | #S | Image Size | Protocol | Ref |
|---|---|---|---|---|---|---|
| CH | Chinese hamster ovary cells | 5 | 327 | 512×382 | 5CV | [35] |
| HE | 2D HELA | 10 | 862 | 512×382 | 5CV | [35] |
| LO | Locate Endogenous | 10 | 502 | 768×512 | 5CV | [36] |
| TR | Locate Transfected | 11 | 553 | 768×512 | 5CV | [36] |
| RN | Fly Cell | 10 | 200 | 1024×1024 | 5CV | [37] |
| TB | Terminal bulb aging | 7 | 970 | 768×512 | 5CV | [37] |
| LY | Lymphoma | 3 | 375 | 1388×1040 | 5CV | [37] |
| MA | Muscle aging | 4 | 237 | 1600×1200 | 5CV | [37] |
| LG | Liver gender | 2 | 265 | 1388×1040 | 5CV | [37] |
| LA | Liver aging | 4 | 529 | 1388×1040 | 5CV | [37] |
| CO | Human colorectal cancer | 8 | 5000 | 150×150 | 10CV | [38] |
| BGR | Breast grading carcinoma | 3 | 300 | 1280×960 | 5CV | [39] |
| LAR | Laryngeal dataset | 4 | 1320 | 1280×960 | Tr-Te | [34] |

Performance evaluation and comparison among all the proposed model and ensembles is performed according image classification accuracy, i.e. the rate of correct classifications; moreover the superiority of a method over another is validated according to the Wilcoxon signed rank test [40].

The methods for pattern classification consist in ensembles of CNN models stochastically designed from a starting architecture by replacing the activation layers. In this section we describe how we derive ensembles of new CNNs starting from a base architecture – this is selected among the best-performing general-purpose networks for image classification, like AlexNet [29], GoogleNet [30],VGGNet [32], ResNet [3], DenseNet [33]).

Starting from the above cited base architectures several stochastic ensembles are creating by the fusion of N stochastic models obtained using the pseudo-code reported in Figure 1. **GenStochasticModel** takes as input a CNN base model IM, a set of activation functions AS and return a new model obtained by randomly replacing all the activation layers of the input model IM with activation layers randomly drawn from AS. For the ensemble creation, first all the N stochastic models are fine-tuned on the training set, then they are fused together in an ensemble using the sum rule, i.e. summing the outputs of their last softmax layer, finally, the final decision is obtained applying an argmax function.

**Figure 1**. Pseudo code of the algorithm for the creation of a stand-alone stochastic model

---

**Function** GenStochasticModel
**Input:**
  Base model: IM
  Set of activation functions AS
**Output:**
  Output model: OM
**Algorithm:**
  Let L be the set of layers of IM
  For each activation $l \in$ L
      Randomly draw an activation function $l'$ from AS
      Replace $l$ with $l'$
**End**

---

In this work we have selected as "backbone" network the ResNet50 [3], which is composed by 50 layers and is one of the most widely architectures for image classification. (anyway the reported experiments could also be replicated for different architectures). The original ResNet50 architecture, which contains ReLu layers to be substituted by a different activation functions, is coupled with three different sets of activation functions to be used as input in the **GenStochasticModel** procedure. The first set, denoted as OldAS, includes the 9 activation functions proposed in [26]: MeLU(k=8), leakyReLU, ELU, MeLU(k=4), PReLU, SReLU, APLU, GaLU, sGaLU. The second set, denoted as FullAS, includes the whole set of activation functions described in section 3, i.e. the same 9 functions of OldAS and a further set of 7 activation functions: ReLU, SoftLearnable, PDeLU, learnableMish, SRS, SwishLearnable, Swish. Finally, in order to evaluate the effectiveness of the new proposed activation functions, a third set is built, named BaseAS, excluding from FullAS all the activation functions proposed by the authors of this paper; this last set is a baseline to our work and includes the following 11 functions: leakyReLU, ELU, PReLU, SReLU, APLU, ReLU, PDeLU, learnableMish, SRS, SwishLearnable, Swish.

The fine tuning of the stochastic CNN models to each image classification problem has been performed according the following training option: batch size 30, max epoch 30, learning rate 0.0001 (for all the network, no freezing); data augmentation includes image reflections on the two axes and random rescaling using a factor uniformly sampled in [1, 2].

## 5. Results

In this section we compare the classification performance of the stochastic ensembles proposed in the previous section with several stand-alone CNN and ensembles of classifiers. The first experiment is aimed at comparing the different activation functions presented in section 3 and the stochastic method for the design on a new model. In table 2 we compare all the variants of ResNet50 architecture obtained by deterministically substituting each activation layer by one of the activation functions of section 3 (the same function for all the network): such CNNs are denoted with the name of the activation function. Some activation functions depend on the training parameter *maxInput* which has been set to 1 or 255 (as reported in parentheses). Moreover the last rows in table2 report the performance of some stochastic models obtained by the calling the **GenStochasticModel** method with ResNet50 and the 3 different activation sets detailed in section 4: the resulting models (they are stand-alone methods) are denoted as SOldAs, SFullAS and SBaseAS, respectively.

**Table 2**. Performance of the several ResNet50 variants in the medical image datasets (accuracy): the last two columns report the average accuracy (Avg) and the rank (evaluated on Avg).

| Method | Dataset | | | | | | | | | | | | | Avg | Rank |
|---|---|---|---|---|---|---|---|---|---|---|---|---|---|---|---|
| | CH | HE | LO | TR | RN | TB | LY | MA | LG | LA | CO | BG | LAR | | |
| ReLU | 93.5 | 89.9 | 95.6 | 90.0 | 55.0 | 58.5 | 77.9 | 90.0 | 93.0 | 85.1 | 94.9 | 88.7 | 87.1 | 84.55 | 6 |
| leakyReLU | 89.2 | 87.1 | 92.8 | 84.2 | 34.0 | 57.1 | 70.9 | 79.2 | 93.7 | 82.5 | 95.7 | 90.3 | 87.3 | 80.30 | 22 |
| ELU | 90.2 | 86.7 | 94.0 | 85.8 | 48.0 | 60.8 | 65.3 | 85.0 | 96.0 | 90.1 | 95.1 | 89.3 | 89.9 | 82.80 | 20 |
| SReLU | 91.4 | 85.6 | 92.6 | 83.3 | 30.0 | 55.9 | 69.3 | 75.0 | 88.0 | 82.1 | 95.7 | 89.0 | 89.5 | 79.02 | 24 |
| APLU | 92.3 | 87.1 | 93.2 | 80.9 | 25.0 | 54.1 | 67.2 | 76.7 | 93.0 | 82.7 | 95.5 | 90.3 | 88.9 | 78.99 | 25 |
| GaLU | 92.9 | 88.4 | 92.2 | 90.4 | 41.5 | 57.8 | 73.6 | 89.2 | 92.7 | 88.8 | 94.9 | 90.3 | 90.0 | 83.28 | 17 |
| sGaLU | 92.3 | 87.9 | 93.2 | 91.1 | 52.0 | 60.0 | 72.5 | 90.0 | 95.3 | 87.4 | 95.4 | 87.7 | 88.8 | 84.13 | 8 |
| PReLU | 92.0 | 85.4 | 91.4 | 81.6 | 33.5 | 57.1 | 68.8 | 76.3 | 88.3 | 82.1 | 95.7 | 88.7 | 89.6 | 79.26 | 23 |
| MeLU | 91.1 | 85.4 | 92.8 | 84.9 | 27.5 | 55.4 | 68.5 | 77.1 | 90.0 | 79.4 | 95.3 | 89.3 | 87.2 | 78.76 | 27 |
| wMeLU | 92.9 | 86.4 | 91.8 | 82.9 | 25.5 | 56.3 | 67.5 | 76.3 | 91.0 | 82.5 | 94.8 | 89.7 | 88.8 | 78.95 | 26 |
| softLearnable2 | 93.9 | 87.3 | 93.6 | 92.5 | 46.0 | 60.3 | 69.0 | 89.5 | 94.6 | 86.1 | 95.0 | 89.6 | 87.0 | 83.41 | 15 |
| softLearnable | 94.1 | 87.4 | 93.4 | 90.3 | 47.0 | 59.1 | 67.7 | 88.3 | 95.0 | 85.5 | 95.5 | 89.3 | 88.2 | 83.13 | 19 |
| pdeluLayer | 94.1 | 87.2 | 92.0 | 91.6 | 51.5 | 56.7 | 70.9 | 89.5 | 96.3 | 86.6 | 95.0 | 89.6 | 88.1 | 83.77 | 12 |
| learnableMishLayer | 95.0 | 87.5 | 93.2 | 91.8 | 45.0 | 58.4 | 69.0 | 86.6 | 95.3 | 86.6 | 95.4 | 90.0 | 88.4 | 83.24 | 18 |
| SRSLayer | 93.2 | 88.8 | 93.4 | 91.0 | 51.5 | 60.1 | 69.8 | 88.7 | 95.0 | 86.4 | 95.7 | 88.3 | 89.4 | 83.94 | 10 |
| swishLearnable | 93.5 | 87.9 | 94.4 | 91.6 | 48.0 | 59.2 | 69.3 | 88.7 | 95.3 | 83.2 | 96.1 | 90.0 | 89.3 | 83.57 | 14 |
| swishLayer | 94.1 | 88.0 | 94.2 | 90.7 | 48.5 | 59.9 | 70.1 | 89.1 | 92.6 | 86.1 | 95.6 | 87.6 | 87.6 | 83.39 | 16 |
| SReLU(255) | 92.3 | 89.4 | 93.0 | 90.7 | 56.5 | 59.7 | 73.3 | 91.7 | 98.3 | 89.0 | 95.5 | 89.7 | 87.9 | 85.15 | 4 |
| APLU(255) | 95.1 | 89.2 | 93.6 | 90.7 | 47.5 | 56.9 | 75.2 | 89.2 | 97.3 | 87.1 | 95.7 | 89.7 | 89.5 | 84.35 | 7 |
| GaLU(255) | 92.9 | 87.2 | 92.0 | 91.3 | 47.5 | 60.1 | 74.1 | 87.9 | 96.0 | 86.9 | 95.6 | 89.3 | 87.7 | 83.73 | 13 |
| sGaLU(255) | 93.5 | 87.8 | 95.6 | 89.8 | 55.0 | 63.1 | 76.0 | 90.4 | 95.0 | 85.3 | 95.1 | 89.7 | 89.8 | 85.09 | 5 |
| MeLU(255) | 92.9 | 90.2 | 95.0 | 91.8 | 57.0 | 59.8 | 78.4 | 87.5 | 97.3 | 85.1 | 95.7 | 89.3 | 88.3 | 85.26 | 2 |
| wMeLU(255) | 94.5 | 89.3 | 94.2 | 92.2 | 54.0 | 61.9 | 75.7 | 89.2 | 97.0 | 88.6 | 95.6 | 87.7 | 88.7 | 85.27 | 1 |
| SOldAS | 90.2 | 90.0 | 94.2 | 91.6 | 54.5 | 62.0 | 77.3 | 90.8 | 95.7 | 90.5 | 95.1 | 89.0 | 87.1 | 85.23 | 3 |
| SOldAS(255) | 93.2 | 88.5 | 94.4 | 91.6 | 51.5 | 59.1 | 73.9 | 88.3 | 94.0 | 89.1 | 95.1 | 86.7 | 88.0 | 84.11 | 9 |
| SFullAS(255) | 94.1 | 87.2 | 93.0 | 87.3 | 54.5 | 60.1 | 72.3 | 89.2 | 94.7 | 83.6 | 94.6 | 89.0 | 89.9 | 83.80 | 11 |
| SBaseAS | 92.6 | 88.0 | 92.4 | 93.1 | 55.5 | 56.8 | 71.7 | 80.4 | 86.7 | 87.2 | 94.6 | 87.0 | 88.5 | 82.65 | 21 |

In the second experiment, reported in Table 3, we compare the performance of several ensembles obtained by the fusion of the previous approaches:
- *FusOldAS10/FusOldAS10(255)* are the fusion by the sum rule of the above models whose activation function belong to the set OldAS; as denoted by their name such ensembles are made of 10 classifiers.
- *FusFullAS16(255)* is the fusion by the sum rule of the 16 above models whose activation function belong to the set FullAS.
- *Sto<setName><K>(maxInput)* denotes the ensembles of stochastic models, i.e. *StoOldAS10* and *StoOldAS10*(255) are ensembles obtained by the fusion of K=10 SOldAS stochastic models; we test ensembles of 5, 10, 15 stochastic models in order to study the dependence of the performance on the number models.
- *FusReLu<K>* denotes the ensembles of K standard ResNet50 models (with ReLu activation functions): it is used to demonstrate the usefulness of stochastic selection of the activation functions to increase diversity in the creation of ensembles.

- *FusOldAS3* and *StoOLDAS3* are the fusion of 3 model taken respectively from the best fixed activation models and 3 stochastic models.
- *A+B* means sum rule between A and B.

**Table 3**. Performance of the proposed ensembles in the medical image datasets (accuracy); the last two columns report the average accuracy (Avg) and the rank (evaluated on Avg).

| Method | Dataset | | | | | | | | | | | | | Avg | Rank |
| --- | --- | --- | --- | --- | --- | --- | --- | --- | --- | --- | --- | --- | --- | --- | --- |
| | CH | HE | LO | TR | RN | TB | LY | MA | LG | LA | CO | BG | LAR | | |
| FusOldAS10 | 93.5 | 90.7 | 97.2 | 92.7 | 56.0 | 63.9 | 77.6 | 90.8 | 96.3 | 91.4 | 96.4 | 90.0 | 90.0 | 86.67 | 20 |
| FusOldAS10(255) | 95.1 | 91.3 | 96.2 | 94.2 | 63.0 | 64.9 | 78.7 | 92.5 | 97.7 | 87.6 | 96.5 | 89.7 | 89.8 | 87.46 | 13 |
| FusFullAS16(255) | 97.2 | 91.3 | 97.4 | 95.5 | 60.0 | 64.5 | 76.0 | 94.2 | 98.3 | 89.1 | 96.8 | 90.0 | 90.3 | 87.74 | 12 |
| StoOldAS10 | 95.4 | 91.3 | 95.8 | 95.1 | 63.0 | 64.2 | 78.9 | 93.8 | 98.7 | 91.1 | 96.5 | 90.3 | 90.2 | 88.02 | 10 |
| StoOldAS(255) | 96.6 | 90.8 | 97.0 | 96.0 | 55.5 | 65.1 | 78.1 | 92.1 | 98.3 | 90.1 | 96.3 | 88.7 | 90.0 | 87.27 | 14 |
| StoOldAS10(255) | 96.9 | 91.2 | 96.8 | 96.2 | 58.5 | 66.6 | 79.7 | 92.5 | 98.3 | 91.6 | 96.6 | 89.7 | 91.1 | 88.13 | 9 |
| StoOldAS15(255) | 97.8 | 91.5 | 96.6 | 95.8 | 60.0 | 65.8 | 80.0 | 92.9 | 99.0 | 91.2 | 96.6 | 90.7 | 91.0 | 88.37 | 8 |
| StoFullAS5(255) | 98.1 | 92.3 | 96.6 | 95.5 | 64.0 | 64.6 | 83.2 | 93.8 | 99.0 | 92.6 | 96.6 | 91.3 | 92.1 | 89.20 | 6 |
| StoFullAS10(255) | 98.8 | 92.9 | 97.6 | 95.8 | 66.5 | 65.7 | 84.3 | 93.7 | 99.3 | 94.1 | 96.8 | 90.3 | 92.3 | 89.85 | 4 |
| StoFullAS15(255) | 98.8 | 93.4 | 97.8 | 96.4 | 65.5 | 66.9 | 85.6 | 92.9 | 99.7 | 94.3 | 96.6 | 91.3 | 92.3 | 90.11 | 2 |
| StoBaseAS5(255) | 99.4 | 91.0 | 97.6 | 95.6 | 61.5 | 64.1 | 80.0 | 88.3 | 94.7 | 91.8 | 96.5 | 90.3 | 90.4 | 87.78 | 11 |
| StoBaseAS10(255) | 99.4 | 93.5 | 97.8 | 95.6 | 65.5 | 65.8 | 81.3 | 89.6 | 96.3 | 94.9 | 96.7 | 91.0 | 90.8 | 89.09 | 7 |
| StoBaseAS15(255) | 99.4 | 93.9 | 98.0 | 96.0 | 64.5 | 66.4 | 83.2 | 90.0 | 96.0 | 93.9 | 96.7 | 92.0 | 91.3 | 89.33 | 5 |
| StoBaseAS8(255)+ StoFullAS7(255) | 99.4 | 93.8 | 98.0 | 96.0 | 67.5 | 66.3 | 83.5 | 92.1 | 98.0 | 95.1 | 96.7 | 91.7 | 91.7 | 89.98 | 3 |
| StoBaseAS15(255)+ StoFullAS15(255) | 99.4 | 93.8 | 98.0 | 96.5 | 67.5 | 67.0 | 85.9 | 91.2 | 98.7 | 94.9 | 96.9 | 92.0 | 92.3 | 90.31 | 1 |
| FusOldAS3(255) | 93.9 | 91.5 | 94.8 | 93.1 | 58.5 | 63.5 | 77.6 | 91.3 | 98.3 | 88.0 | 96.3 | 89.0 | 89.4 | 86.55 | 21 |
| StoOldAS3(255) | 96.3 | 90.9 | 95.6 | 95.1 | 54.0 | 62.9 | 78.7 | 92.5 | 98.7 | 90.9 | 96.2 | 90.0 | 90.5 | 87.10 | 16 |
| FusRelu5 | 95.0 | 90.5 | 96.2 | 94.7 | 56.0 | 63.7 | 77.1 | 94.1 | 95.6 | 89.1 | 96.4 | 89.0 | 89.5 | 86.68 | 19 |
| FusRelu10 | 94.5 | 91.6 | 95.8 | 94.5 | 56.5 | 64.5 | 76.0 | 93.3 | 97.7 | 89.1 | 96.6 | 89.6 | 90.2 | 86.91 | 17 |
| FusRelu15 | 95.4 | 91.1 | 96.2 | 95.1 | 58.5 | 64.8 | 76.0 | 92.9 | 97.3 | 89.3 | 96.3 | 90.0 | 90.4 | 87.17 | 15 |
| FusRelu30 | 95.4 | 91.5 | 96.2 | 94.7 | 59.0 | 63.9 | 75.7 | 92.5 | 97.3 | 88.2 | 96.4 | 89.0 | 90.1 | 86.91 | 18 |

We shall now summarize the most relevant results reported in Tables 2 and 3:
- Ensemble methods outperform stand-alone networks. Hence, changing the activation functions is an effective way to create ensembles of networks.
- Among stand-alone networks, ReLU is not the best one. The activations that reach the highest performance are the two MeLUs with *maxInput* equal to 255. This results is achieved although the ResNet architecture was originally designed for ReLU.
- The performance of the networks that use different activations in the same architecture is not consistently better than the other stand-alone networks, but they show their power when are combined to create an ensemble. When used together, they reach the best overall results.
- Considering the ensembles, *FusReLu<K>* which is an ensemble of K standard ResNet50 is a baseline for ensemble approaches; it performs well due to instability of tuning in small datasets but it is outperformed by almost all the stochastic approaches (*Sto<setName><K>(maxInput)*).
- The four *FusReLu<K>* ensembles have similar performance (p-value >0.1), but any of them outperforms stand-alone ReLU (p-value 0.005)
- Considering the number of models in each ensemble, we can see that increasing K from 5 to 10 to 15 and to 30 the ensembles seem to perform better, anyway there is no statistical evidence of a difference from 15 and 30.
- The non-stochastic ensemble *FusFullAS16*(255) outperforms with a p-value 0.05 the ensemble of ReseNet50 *FusReLu15*; instead *FusOldAS10(255)* obtains performance statistically similar to

*FusReLu10* and *FusReLu15*. Anyway the stochastic ensemble *StoOldAS15(255)* outperforms with a p-value of 0.1 *FusFullAS16(255)*.

- Designing the model by means of stochastic activation functions gives valuable results, especially in the creation of ensembles. Indeed, these approaches are the first ranked methods tested in these experiments.
- The comparison between the two "light" ensembles, *FusOldAS3* and *StoOLDAS3,* which are made of fixed and stochastic models, respectively, suggests again that using stochastic activation functions improves the performance in ensembles.
- According to the Wilcoxon Signed Rank Test *StoFullAS15*(255) outperforms *StoOldAS15*(255)FusRanOLD15 with a p-value 0.0001 and *StoBaseAS15*(255) with a p-value of 0.1, instead there is not statist difference between *StoFullAS15*(255) vs *StoBaseAS8(255)+StoFullAS7(255)* and *StoFullAS15*(255) vs *StoFullAS15*(255) + *StoBaselAS15*(255)

In our experiments, we used a GTX1080 GPU. ResNet50 can classify more than 40 images; hence, a 20 networks ensemble can classify two images per second using a single GTX1080.

**4. Conclusions**

In this paper, we presented three different methods to modify an existing CNN architecture to obtain new high performance and diverse networks. The core idea of our methods is replacing the activation functions in the original network with other ones, which can be different in every layer. This process allowed us to create new networks well suited to be included in an ensemble. We showed that our methods outperform standard ResNet50, all our stand-alone methods and all our ensemble baselines on a variety of biomedical datasets. In this paper, we only used ResNet50 as backbone architecture, but we plan to test lighter networks suitable for mobile devices as future work. Studying ensembles of CNNs requires large computational power and long, memory expensive experiments.

All the code will be available at https://github.com/LorisNanni in order to encourage the replication of our work.

**Acknowledgments:** We gratefully acknowledge the support of NVIDIA Corporation for the "NVIDIA Hardware Donation Grant" of a Titan X used in this research.

**Conflicts of Interest:** The authors declare no conflict of interest.

**References**

1. Goodfellow, I.; Bengio, Y.; Courville, A. *Deep Learning*; MIT Press, 2016;

2. Cho, Y.; Saul, L.K. Large-margin classification in infinite neural networks. *Neural Comput*. 2010.

3. He, K.; Zhang, X.; Ren, S.; Sun, J. Deep Residual Learning for Image Recognition. In Proceedings of the 2016 IEEE Conference on Computer Vision and Pattern Recognition (CVPR); 2016; pp. 770–778.

4. Hansen, L.K.; Salamon, P. Neural Network Ensembles. *IEEE Trans. Pattern Anal. Mach. Intell.* **1990**, doi:10.1109/34.58871.

5. Ghosh, P.; Antani, S.; Long, L.R.; Thoma, G.R. Review of medical image retrieval systems and future directions. In Proceedings of the Proceedings - IEEE Symposium on Computer-Based Medical Systems; 2011.

6. Esteva, A.; Kuprel, B.; Novoa, R.A.; Ko, J.; Swetter, S.M.; Blau, H.M.; Thrun, S. Dermatologist-level classification of skin cancer with deep neural networks. *Nature* **2017**, *542*, 115–118,

doi:10.1038/nature21056.

7. Paci, M.; Nanni, L.; Lahti, A.; Aalto-Setala, K.; Hyttinen, J.; Severi, S. Non-Binary Coding for Texture Descriptors in Sub-Cellular and Stem Cell Image Classification. *Curr. Bioinform.* **2013**, doi:10.2174/1574893611308020009.

8. Chi, J.; Walia, E.; Babyn, P.; Wang, J.; Groot, G.; Eramian, M. Thyroid nodule classification in ultrasound images by fine-tuning deep convolutional neural network. *J. Digit. Imaging* **2017**, *30*, 477–486.

9. Byra, M. Discriminant analysis of neural style representations for breast lesion classification in ultrasound. *Biocybern. Biomed. Eng.* **2018**, *38*, 684–690.

10. Nanni, L.; Brahnam, S.; Ghidoni, S.; Lumini, A. Bioimage Classification with Handcrafted and Learned Features. *IEEE/ACM Trans. Comput. Biol. Bioinforma.* **2018**, doi:10.1109/TCBB.2018.2821127.

11. Alom, M.Z.; Taha, T.M.; Yakopcic, C.; Westberg, S.; Sidike, P.; Nasrin, M.S.; Hasan, M.; Van Essen, B.C.; Awwal, A.A.S.; Asari, V.K. A State-of-the-Art Survey on Deep Learning Theory and Architectures. *Electronics* **2019**, *8*, 292, doi:10.3390/electronics8030292.

12. Liu, W.; Wang, Z.; Liu, X.; Zeng, N.; Liu, Y.; Alsaadi, F.E. A survey of deep neural network architectures and their applications. *Neurocomputing* **2017**, doi:10.1016/j.neucom.2016.12.038.

13. Agostinelli, F.; Hoffman, M.; Sadowski, P.; Baldi, P. Learning activation functions to improve deep neural networks. In Proceedings of the 3rd International Conference on Learning Representations, ICLR 2015 - Workshop Track Proceedings; 2015.

14. Clevert, D.A.; Unterthiner, T.; Hochreiter, S. Fast and accurate deep network learning by exponential linear units (ELUs). In Proceedings of the 4th International Conference on Learning Representations, ICLR 2016 - Conference Track Proceedings; 2016.

15. Glorot, X.; Bordes, A.; Bengio, Y. Deep sparse rectifier neural networks. In Proceedings of the Journal of Machine Learning Research; 2011.

16. Maas, A.L.; Hannun, A.Y.; Ng, A.Y. Rectifier nonlinearities improve neural network acoustic models. In Proceedings of the in ICML Workshop on Deep Learning for Audio, Speech and Language Processing; 2013.

17. Klambauer, G.; Unterthiner, T.; Mayr, A.; Hochreiter, S. Self-Normalizing Neural Networks. In Proceedings of the NIPS; 2017.

18. Xu, B.; Wang, N.; Chen, T.; Li, M. Empirical Evaluation of Rectified Activations in Convolutional Network 2015.

19. He, K.; Zhang, X.; Ren, S.; Sun, J. Delving deep into rectifiers: Surpassing human-level performance on imagenet classification. *Proc. IEEE Int. Conf. Comput. Vis.* **2015**, *2015 Inter*, 1026–1034, doi:10.1109/ICCV.2015.123.

20. Ertuğrul, Ö.F. A novel type of activation function in artificial neural networks: Trained activation function. *Neural Networks* **2018**, *99*, 148–157, doi:10.1016/J.NEUNET.2018.01.007.

21. Bawa, V.S.; Kumar, V. Linearized sigmoidal activation: A novel activation function with tractable non-linear characteristics to boost representation capability. *Expert Syst. Appl.* **2019**, *120*, 346–356, doi:10.1016/J.ESWA.2018.11.042.

22. Ramachandran, P.; Barret, Z.; Le, Q. V. Searching for activation functions. In Proceedings of the 6th International Conference on Learning Representations, ICLR 2018 - Workshop Track Proceedings; 2018.

23. Maguolo, G.; Nanni, L.; Ghidoni, S. Ensemble of Convolutional Neural Networks Trained with Different Activation Functions. **2019**.

24. He, K.; Zhang, X.; Ren, S.; Sun, J. Delving deep into rectifiers: Surpassing human-level performance on imagenet classification. In Proceedings of the Proceedings of the IEEE International Conference on Computer Vision; 2015.

25. Jin, X.; Xu, C.; Feng, J.; Wei, Y.; Xiong, J.; Yan, S. Deep learning with S-shaped rectified linear activation units. In Proceedings of the 30th AAAI Conference on Artificial Intelligence, AAAI 2016; 2016.

26. Nanni, L.; Lumini, A.; Ghidoni, S.; Maguolo, G. Stochastic selection of activation layers for convolutional neural networks. *Sensors (Switzerland)* **2020**, doi:10.3390/s20061626.

27. Misra, D. Mish: A Self Regularized Non-Monotonic Activation Function.

28. Zhou, Y.; Li, D.; Huo, S.; Kung, S.-Y. Soft-Root-Sign Activation Function. **2020**.

29. Krizhevsky, A.; Sutskever, I.; Hinton, G.E. ImageNet Classification with Deep Convolutional Neural Networks. *Adv. Neural Inf. Process. Syst.* **2012**, 1–9, doi:http://dx.doi.org/10.1016/j.protcy.2014.09.007.

30. Szegedy, C.; Liu, W.; Jia, Y.; Sermanet, P.; Reed, S.; Anguelov, D.; Erhan, D.; Vanhoucke, V.; Rabinovich, A. Going deeper with convolutions. In Proceedings of the Proceedings of the IEEE Computer Society Conference on Computer Vision and Pattern Recognition; 2015; Vol. 07-12-June, pp. 1–9.

31. Szegedy, C.; Vanhoucke, V.; Ioffe, S.; Shlens, J.; Wojna, Z. Rethinking the Inception Architecture for Computer Vision. In Proceedings of the 2016 IEEE Conference on Computer Vision and Pattern Recognition (CVPR); 2016; Vol. 00, pp. 2818–2826.

32. Simonyan, K.; Zisserman, A. Very Deep Convolutional Networks for Large-Scale Image Recognition. *Int. Conf. Learn. Represent.* **2015**, 1–14, doi:10.1016/j.infsof.2008.09.005.

33. Huang, G.; Liu, Z.; Van Der Maaten, L.; Weinberger, K.Q. Densely connected convolutional networks. In Proceedings of the Proceedings - 30th IEEE Conference on Computer Vision and Pattern Recognition, CVPR 2017; 2017.

34. Moccia, S.; De Momi, E.; Guarnaschelli, M.; Savazzi, M.; Laborai, A.; Guastini, L.; Peretti, G.; Mattos, L.S. Confident texture-based laryngeal tissue classification for early stage diagnosis support. *J. Med. Imaging*

**2017**, *4*, 34502.

35. Boland, M. V; Murphy, R.F. A neural network classifier capable of recognizing the patterns of all major subcellular structures in fluorescence microscope images of HeLa cells. *Bioinformatics* **2001**, *17*, 1213–1223, doi:10.1093/bioinformatics/17.12.1213.

36. Hamilton, N.A.; Pantelic, R.S.; Hanson, K.; Teasdale, R.D. Fast automated cell phenotype image classification. *BMC Bioinformatics* **2007**, *8*, 110, doi:10.1186/1471-2105-8-110.

37. Shamir, L.; Orlov, N.; Mark Eckley, D.; Macura, T.J.; Goldberg, I.G. IICBU 2008: A proposed benchmark suite for biological image analysis. *Med. Biol. Eng. Comput.* **2008**, *46*, 943–947, doi:10.1007/s11517-008-0380-5.

38. Kather, J.N.; Weis, C.-A.; Bianconi, F.; Melchers, S.M.; Schad, L.R.; Gaiser, T.; Marx, A.; Zöllner, F.G. Multi-class texture analysis in colorectal cancer histology. *Sci. Rep.* **2016**, *6*, 27988, doi:10.1038/srep27988.

39. Dimitropoulos, K.; Barmpoutis, P.; Zioga, C.; Kamas, A.; Patsiaoura, K.; Grammalidis, N. Grading of invasive breast carcinoma through Grassmannian VLAD encoding. *PLoS One* **2017**, *12*, 1–18, doi:10.1371/journal.pone.0185110.

40. Demšar, J. Statistical Comparisons of Classifiers over Multiple Data Sets. *J. Mach. Learn. Res.* **2006**, *7*, 1–30, doi:10.1016/j.jecp.2010.03.005.